# Evolutionary Cost-sensitive Extreme Learning Machine

Lei Zhang, *Member, IEEE*, and David Zhang, *Fellow, IEEE*

*Abstract*—Conventional extreme learning machines solve a Moore-Penrose generalized inverse of hidden layer activated matrix and analytically determine the output weights to achieve generalized performance, by assuming the same loss from different types of misclassification. The assumption may not hold in cost-sensitive recognition tasks, such as face recognition based access control system, where misclassifying a stranger as a family member may result in more serious disaster than misclassifying a family member as a stranger. Though recent cost-sensitive learning can reduce the total loss with a given cost matrix that quantifies how severe one type of mistake against another, in many realistic cases the cost matrix is unknown to users. Motivated by these concerns, this paper proposes an evolutionary cost-sensitive extreme learning machine (ECSELM), with the following merits: 1) to our best knowledge, it is the first proposal of ELM in evolutionary cost-sensitive classification scenario; 2) it well addresses the open issue of how to define the cost matrix in cost-sensitive learning tasks; 3) an evolutionary backtracking search algorithm is induced for adaptive cost matrix optimization. Experiments in a variety of cost-sensitive tasks well demonstrate the effectiveness of the proposed approaches, with about 5%~10% improvements.

## I. INTRODUCTION

Extreme learning machine (ELM) was proposed by Huang [1] for generalized single-hidden-layer feed-forward neural networks (SLFN) in order to overcome the drawbacks of gradient-based methods, such as the local minima, learning rate, stopping criteria and learning epochs. As Huang, et al has further provided the rigorous proof of universal approximation of ELM with much milder condition that almost any nonlinear piecewise continuous function can be used as the activation functions in hidden nodes of ELM [1]-[4]. Different from traditional learning algorithms, ELM not only tends to reach the smallest training error but also the smallest norm of the output weights for better generalization performance of SLFN, according to the Bartlett's theory [5]. The most recent advances of ELM about its biological understanding and fast deep learning perspectives can be found in [6]-[9].

ELM, in which the input weights and hidden biases were randomly selected and the output weights were analytically determined using Moore-Penrose generalized inverse, has also been proved to be efficient and effective for regression and classification tasks [10]-[14]. An excellent review of ELM can refer to as [15]. However, ELM may require more hidden neurons than gradient descent algorithms due to the randomly selected input weights and hidden biases [16].

Different versions of improved ELM have been proposed. Inspired by Mercer condition, a kernel ELM was proposed for robust classification [10]. Under kernels, a sequential ELM approach [17] was also proposed for online learning by using the kernel recursive least-squares, an extension of kernel adaptive filtering. Also, a recursive orthogonal least-square method combined with sequential partial orthogonalization was incorporated into ELM, which formulates a new parsimonious ELM [18] and has been used for nonlinear time-series modeling. Considering that the dense weights of ELM easily lead to overfitting, a sparse Bayesian ELM [19] was proposed to improve the robustness by pruning the redundant hidden neurons in learning phase, such that the model is insensitive to hidden neurons. It is known that in ELM, the hidden nodes are generally frozen such that the learning ability may be limited. Therefore, an ELM with adaptive growth of hidden nodes was proposed in [20], and achieved automated design of networks. It was also verified that more compact network architecture can be achieved. Since ELM randomly selects the input weights and biases for feature mapping, in [16], a differential evolutionary based ELM (E-ELM) was proposed to optimize the random input weights and tend to improve the generalization performance with compact networks. Though 'evolutionary' concept is also used, essential difference between the proposed evolutionary cost-sensitive extreme learning machine (ECSELM) and E-ELM is witnessed. Specifically, our proposed ECSELM is to conduct an optimal cost-sensitive learning for handling the same-loss problem supposed in ELM, but not aim at optimizing the random weights and bias addressed in E-ELM. For tackling a recognition task with imbalanced datasets that are quite common in various applications, a weighted ELM (WELM) [21] is proposed, where each training sample was assigned with larger weight to strengthen the impact of minority class and smaller weight to weak the impact of majority class. Further, a boosting weighted ELM was also proposed with an AdaBoost framework for sample imbalance [22], [23] and a cost-sensitive ELM (CSELM) [24] was proposed for sample imbalance

L. Zhang is with the College of Communication Engineering, Chongqing University, China (e-mail: leizhang@cqu.edu.cn).
D. Zhang is with the Department of Computing, The Hong Kong Polytechnic University, Hong Kong (e-mail: csdzhang@comp.polyu.edu.hk).

weighting. ELM, due to its efficacy, has drawn a significant amount of interest from researchers in various fields, such as face recognition [12], [25], activity recognition [26], action recognition [27], and handwritten character recognition [28].

Up to now, ELM with many variants has been widely used for classification and regression. However, all the existing ELM based recognition methods tend to achieve lower error rate by supposing the same loss for any misclassification, which, however, may not hold in many applications, for instance, face recognition based access control system, as different mistakes may lead to different losses. Specifically, it would be a serious disaster if the system misclassifies a stranger as a family member and allowed to enter the room. Instead, misclassifying a family member as a stranger and not allowed to enter the room may be less serious. The different losses in a face recognition system have been first paid an attention by formulating a cost-sensitive classification task [29].

Subspace methods such as principal component analysis (PCA) [30], linear discriminant analysis (LDA) [31], manifold learning based locality preserving projections (LPP) [32], margin fisher analysis (MFA) [33], and their kernelized, and tensorized variants [34], [35] have been proposed. Recently, their cost-sensitive variants, such as CSPCA, CSLDA, CSLPP, and CSMFA have also been surveyed for face recognition in [36], [37]. Cost sensitive learning can reduce the loss with a predefined cost matrix that quantifies how severe one type of mistake against another type of mistake, but in many realistic cases the cost matrix is unknown or difficult to define by users [29], such that the learned subspace is not optimal with poor classification performance. Note that misclassification loss is produced due to incorrectly classifying one sample in the $i$-th class as the $j$-th class. In many realistic cases, users only know that one type of mistake is more serious than another type, but it is difficult to specify a cost value of a mistake. In [38] the authors first attempt to address the problem of cost matrix definition using a cost interval (e.g. a possible cost range) instead of a precise cost value, but it brings a large computational cost and the cost interval should be manually pre-defined, such that the cost-matrix definition is still an open topic in cost-sensitive learning. Learning a cost matrix is extremely desired to be resolved for cost-sensitive system. In terms of the final classification task, a good cost matrix should not degrade the recognition accuracy. Therefore, our goal is to optimize the cost matrix for improving the final classification, where the cost-sensitive behavior is modeled.

Motivated by the above open problems of ELM and cost-sensitive learning, an ECSELM is proposed in this paper, which on one hand brings a cost-sensitive ELM with the lowest misclassification loss at the first time, and simultaneously learns an optimal cost matrix automatically on the other hand during CSELM learning. To our best knowledge, this is the first proposal of cost-sensitive ELM as a new perspective. This paper is also a forward-looking work for automatic cost matrix determination. Note that the proposed method has essential difference from that of [24] which does not focus on the cost matrix learning, yet only defines the weights.

The remainder of this paper is as follows. Section II presents the related work of ELMs. The proposed ECSELM and algorithms are formulated in Section III. Experiments on multi-modal dataset for attractiveness prediction are employed in Section IV. Experiments on face datasets for face recognition and verification are conducted in Section V. Experiments on E-NOSE dataset for gases recognition are presented in Section VI. The performance evaluation of classifiers is given in Section VII. The parameter sensitivity analysis is conducted in Section VIII. The complexity analysis is discussed in Section IX. Finally, Section X concludes the paper.

## II. RELATED WORK

### A. SLFN and ELM

Given $N$ samples $[\mathbf{x}_1, \mathbf{x}_2, \cdots, \mathbf{x}_N]$ and their corresponding targets $[\mathbf{t}_1, \mathbf{t}_2, \cdots, \mathbf{t}_N]$, where $\mathbf{x}_i = [x_{i1}, x_{i1}, \cdots, x_{in}]^T \in \mathbb{R}^n$ and $\mathbf{t}_i = [t_{i1}, t_{i1}, \cdots, t_{im}]^T \in \mathbb{R}^m$, standard SLFN with $L$ hidden nodes and activation function $\mathcal{H}(\mathbf{x})$ is modeled as

$$\sum_{j=1}^{L} \boldsymbol{\beta}_j \mathcal{H}(\mathbf{w}_j^T \cdot \mathbf{x}_i + b_j) = \mathbf{t}_i, i = 1, \cdots, N \quad (1)$$

where $\mathbf{w}_j = [w_{j1}, \cdots, w_{jn}]^T$ is the input weight vector connecting the $j$-th hidden node and the $n$ input nodes, $\boldsymbol{\beta}_j = [\beta_{j1}, \cdots, \beta_{jm}]^T$ is the output weight vector connecting the $j$-th hidden node and the $m$ output nodes, $b_j$ is the bias of the $j$-th hidden node. In ELM [1], input weights $\mathbf{w}$ and hidden biases $\mathbf{b}$ are randomly generated independently of the training data.

The representation (1) can be written compactly as

$$\mathbf{H} \cdot \boldsymbol{\beta} = \mathbf{T} \quad (2)$$

where $\boldsymbol{\beta} = [\boldsymbol{\beta}_1, \boldsymbol{\beta}_2, \cdots, \boldsymbol{\beta}_L]^T$, $\mathbf{T} = [\mathbf{t}_1, \mathbf{t}_2, \cdots, \mathbf{t}_N]^T$, $\mathbf{H}_{N \times L}$ is the hidden layer output matrix, and the $i$-th column of $\mathbf{H}$ is the output of the $i$-th hidden neuron $w.r.t$ inputs $\mathbf{x}_1, \mathbf{x}_2, \cdots, \mathbf{x}_N$. Find the *minimum norm least square* solution of the linear system (2) is equivalent to train a SLFN. When the number of hidden neurons $L = N$, $\mathbf{H}$ is a square matrix and invertible. However, in most case the number $L \ll N$, and $\mathbf{H}$ is non-square, therefore, the *minimum norm least square* solution can be solved as

$$\hat{\boldsymbol{\beta}} = \mathbf{H}^{\dagger} \mathbf{T} \quad (3)$$

where $\mathbf{H}^{\dagger}$ is the Moore-Penrose generalized inverse of $\mathbf{H}$.

ELM [1] is to minimize the training error and the $\ell_2$-norm of the output weights, which can be formulated as

$$\text{minimize } \mathcal{L}_{\text{ELM}} = \frac{1}{2} \|\boldsymbol{\beta}\|^2 + C \cdot \frac{1}{2} \cdot \sum_{i=1}^{N} \|\boldsymbol{\xi}_i\|^2 \quad (4)$$
$$\text{subject to: } \mathcal{H}(\mathbf{x}_i)\boldsymbol{\beta} = \mathbf{t}_i - \boldsymbol{\xi}_i, i = 1, \ldots, N$$

where $C$ is the regularization parameter, $\boldsymbol{\xi}_i$ denotes the residual of prediction. As described in [1], [10], by solving problem (4), the output weights can be easily and analytically determined as

$$\boldsymbol{\beta} = \mathbf{H}^{\dagger} \mathbf{T} = \begin{cases} \mathbf{H}^T \left(\frac{\mathbf{I}}{C} + \mathbf{H}\mathbf{H}^T\right)^{-1} \mathbf{T}, N < L \\ \left(\frac{\mathbf{I}}{C} + \mathbf{H}^T\mathbf{H}\right)^{-1} \mathbf{H}^T \mathbf{T}, N \geq L \end{cases} \quad (5)$$

where $\mathbf{I}$ is an identity matrix.

### B. Kernel ELM

One can apply Mercer's condition on ELM and formulate a kernel ELM (KELM) [10]. A kernel in ELM is defined as

$$\boldsymbol{\Omega}_{\text{ELM}} = \mathbf{H}\mathbf{H}^T \quad (6)$$

where $\Omega_{\text{ELM}i,j} = \mathcal{H}(\mathbf{x}_i) \cdot \mathcal{H}(\mathbf{x}_j) = K(\mathbf{x}_i, \mathbf{x}_j)$.

Then, for the case where the number of training samples is not huge (i.e. $N<L$), the output of KELM classifier (6) with respect to the input $\mathbf{x}$, can be represented as

$$y = \mathcal{H}(\mathbf{x})\mathbf{H}^\mathrm{T}\left(\frac{\mathbf{I}}{C} + \mathbf{H}\mathbf{H}^\mathrm{T}\right)^{-1}\mathbf{T}$$
$$= \begin{bmatrix} K(\mathbf{x},\mathbf{x}_1) \\ \vdots \\ K(\mathbf{x},\mathbf{x}_N) \end{bmatrix}^\mathrm{T} \left(\frac{\mathbf{I}}{C} + \mathbf{\Omega}_{\mathrm{ELM}}\right)^{-1}\mathbf{T} \quad (7)$$

### C. Weighted ELM

The weighted ELM was proposed to address the problem of imbalanced samples [21]. In contrast to the ELM, a constant weight matrix $\mathbf{W}$ associated with the number of each class is embedded in the objective function. Therefore, the optimization problem can be rewritten as

$$\text{minimize } \mathcal{L}_{\mathrm{WELM}} = \frac{1}{2}\|\boldsymbol{\beta}\|^2 + C \cdot \mathbf{W} \cdot \frac{1}{2} \cdot \sum_{i=1}^{N}\|\boldsymbol{\xi}_i\|^2 \quad (8)$$
$$\text{subject to: } \mathcal{H}(\mathbf{x}_i)\boldsymbol{\beta} = \mathbf{t}_i - \boldsymbol{\xi}_i, i = 1, \dots, N$$

Generally, each training sample was assigned with larger weight to strength the impact of minority class and smaller weight to weak the majority class. Specifically, two weighted ELM schemes called as $W^1$ELM and $W^2$ELM were given.

$$W^1\text{ELM: } W_{cc} = \frac{1}{\#\text{Class } c} \quad (9)$$

$$W^2\text{ELM: } W_{cc} = \begin{cases} \frac{0.618}{\#\text{Class } c}, & \text{if } \#\text{Class } c > AVG(\#\text{Class } c) \\ \frac{1}{\#\text{Class } c}, & \text{otherwise} \end{cases} \quad (10)$$

where #Class $c$ is the number of samples belonging to class $c$, AVG(#Class $c$) is the average number. Notably 0.618 denotes the golden ratio [21]. New trend on ELM is referred as [39].

## III. THE PROPOSED APPROACHES

### A. Cost-sensitive Extreme Learning Machine

Cost-sensitive learning is an important topic in machine learning. However, cost-sensitive ELM (CSELM) is first proposed as a new perspective for ELMs. In the proposed approach, a cost matrix specifying different costs with respect to different types of misclassification is integrated into the popular ELM, such that the proposed CSELM can be adapted to cost-sensitive learning tasks and scenarios.

The cost matrix $\mathcal{M}$ of $N$ samples can be represented as

$$\mathcal{M} = \begin{bmatrix} 0 & \mathcal{M}_{12} & \cdots & \mathcal{M}_{1q} & \cdots & \mathcal{M}_{1N} \\ \mathcal{M}_{21} & 0 & \cdots & \mathcal{M}_{2q} & \cdots & \mathcal{M}_{2N} \\ \vdots & \vdots & \ddots & \vdots & \cdots & \vdots \\ \mathcal{M}_{q1} & \mathcal{M}_{q2} & \cdots & 0 & \cdots & \mathcal{M}_{qN} \\ \vdots & \vdots & \cdots & \vdots & \ddots & \vdots \\ \mathcal{M}_{N1} & \mathcal{M}_{N2} & \cdots & \mathcal{M}_{Nq} & \cdots & 0 \end{bmatrix}_{N \times N} \quad (11)$$

where $\mathcal{M}_{i,j}$ denotes the misclassification loss that classifies the $i$-th sample as the $j$-th sample, and the diagonal elements of zero denote the correct classification without loss. Then, the proposed CSELM for recognition and regression is shown as

$$\text{minimize } \mathcal{L}_{\mathrm{CSELM}} = \frac{1}{2}\|\boldsymbol{\beta}\|^2 + C \cdot \text{diag}(\boldsymbol{\mathcal{B}}) \cdot \frac{1}{2} \cdot \sum_{i=1}^{N}\|\boldsymbol{\xi}_i\|^2 \quad (12)$$
$$\text{subject to: } \mathcal{H}(\mathbf{x}_i)\boldsymbol{\beta} = \mathbf{t}_i - \boldsymbol{\xi}_i, i = 1, \dots, N$$

where $\boldsymbol{\mathcal{B}}$ is a cost information vector with entries $\mathcal{B}_i = \sum_j (\boldsymbol{\mathcal{W}} \cdot \boldsymbol{\mathcal{M}})_{ij}$, $\boldsymbol{\mathcal{W}}_{N \times N}$ is a diagonal weighted matrix assigned for each training sample whose coefficient can be calculated as (9), such that the cost information vector $\boldsymbol{\mathcal{B}}$ on the error term is also an effective tradeoff between the samples' imbalance and the misclassification loss. Note that there is essential difference between (12) and (8), in that a constant matrix $\mathbf{W}$ in (8) is simply calculated in terms of the sample number, while in (12) we seriously consider the misclassification loss by an unsolved cost information vector $\boldsymbol{\mathcal{B}}$ in cost-sensitive tasks. Therefore, the learning of $\boldsymbol{\mathcal{M}}$ is a key part of CSELM. $\mathbf{t}_i$ and $\boldsymbol{\xi}_i$ denote the label vector and error vector with respect to the sample $\mathbf{x}_i$, for multi-class recognition. If $\mathbf{x}_i$ belongs to the $c$-th class, the $c$-th position of $\mathbf{t}_i$ is set as 1, and -1 otherwise.

With a fixed $\boldsymbol{\mathcal{B}}$, the representation (12) is a convex optimization problem, which can be solved as

$$\mathcal{L}_{\mathrm{CSELM}}(\boldsymbol{\beta},\boldsymbol{\xi}_i,\alpha_i) = \frac{1}{2}\|\boldsymbol{\beta}\|^2 + C \cdot \text{diag}(\boldsymbol{\mathcal{B}}) \cdot \frac{1}{2} \cdot \sum_{i=1}^{N}\|\boldsymbol{\xi}_i\|^2 - \alpha_i \cdot (\mathcal{H}(\mathbf{x}_i)\boldsymbol{\beta} - \mathbf{t}_i + \boldsymbol{\xi}_i) \quad (13)$$

where $\alpha_i$ is the Lagrange multiplier.

To derive the output weights, we calculate the derivatives of $\mathcal{L}_{\mathrm{CSELM}}$ with respect to $\boldsymbol{\beta},\boldsymbol{\xi}_i,\alpha_i$ as follows

$$\begin{cases} \frac{\partial \mathcal{L}(\boldsymbol{\beta},\boldsymbol{\xi}_i,\alpha_i)}{\partial \boldsymbol{\beta}} = 0 \rightarrow \boldsymbol{\beta} = \mathbf{H}^\mathrm{T}\boldsymbol{\alpha} \\ \frac{\partial \mathcal{L}(\boldsymbol{\beta},\boldsymbol{\xi}_i,\alpha_i)}{\partial \boldsymbol{\xi}_i} = 0 \rightarrow \alpha_i = C \cdot \text{diag}(\boldsymbol{\mathcal{B}}) \cdot \boldsymbol{\xi}_i, i = 1,\cdots,N \\ \frac{\partial \mathcal{L}(\boldsymbol{\beta},\boldsymbol{\xi}_i,\alpha_i)}{\partial \alpha_i} = 0 \rightarrow \mathcal{H}(\mathbf{x}_i)\boldsymbol{\beta} - \mathbf{t}_i + \boldsymbol{\xi}_i, i = 1,\cdots,N \end{cases} \quad (14)$$

Then the output weights associated with $\boldsymbol{\mathcal{B}}$ can be solved as
$$\boldsymbol{\beta}_{\boldsymbol{\mathcal{B}}} = \mathbf{H}^\dagger \mathbf{T}$$
$$= \begin{cases} \mathbf{H}^\mathrm{T} \cdot \left(\frac{\mathbf{I}}{C} + \text{diag}(\boldsymbol{\mathcal{B}})\mathbf{H}\mathbf{H}^\mathrm{T}\right)^{-1} \cdot \text{diag}(\boldsymbol{\mathcal{B}}) \cdot \mathbf{T}, N < L \\ \left(\frac{\mathbf{I}}{C} + \mathbf{H}^\mathrm{T}\text{diag}(\boldsymbol{\mathcal{B}})\mathbf{H}\right)^{-1} \cdot \mathbf{H}^\mathrm{T} \cdot \text{diag}(\boldsymbol{\mathcal{B}}) \cdot \mathbf{T}, N \geq L \end{cases} \quad (15)$$

where $\mathbf{H}^\dagger$ is the Moore-Penrose generalized inverse of $\mathbf{H}$, which can be represented as

$$\mathbf{H} = \begin{bmatrix} \mathcal{H}(\mathbf{w}_1\mathbf{x}_1+b_1) & \mathcal{H}(\mathbf{w}_2\mathbf{x}_1+b_2) & \cdots & \mathcal{H}(\mathbf{w}_L\mathbf{x}_1+b_L) \\ \mathcal{H}(\mathbf{w}_1\mathbf{x}_2+b_1) & \mathcal{H}(\mathbf{w}_2\mathbf{x}_2+b_2) & \cdots & \mathcal{H}(\mathbf{w}_L\mathbf{x}_2+b_L) \\ \vdots & \vdots & \vdots & \vdots \\ \mathcal{H}(\mathbf{w}_1\mathbf{x}_N+b_1) & \mathcal{H}(\mathbf{w}_2\mathbf{x}_N+b_2) & \cdots & \mathcal{H}(\mathbf{w}_L\mathbf{x}_N+b_L) \end{bmatrix} \quad (16)$$

In this paper, the "radbas" function is empirically used as the feature mapping (activation) $\mathcal{H}(\cdot)$, which is shown as

$$\mathcal{H}(\mathbf{w},\mathbf{b},\mathbf{x}) = \exp(-\|\mathbf{w}\cdot\mathbf{x}+\mathbf{b}\|^2) \quad (17)$$

Accordingly, sigmoid, Laplacian, polynomial function, etc. can also be used as hidden layer activation function.

The output $\mathbf{z}$ of a test instance $\mathbf{y}$ can be solved with two cases of small sample and huge samples, respectively as
$$\mathbf{z} = \mathcal{H}(\mathbf{y}) \cdot \boldsymbol{\beta}_{\boldsymbol{\mathcal{B}}}$$
$$= \begin{cases} \mathcal{H}(\mathbf{y}) \cdot \mathbf{H}^\mathrm{T} \cdot \left(\frac{\mathbf{I}}{C} + \text{diag}(\boldsymbol{\mathcal{B}})\mathbf{H}\mathbf{H}^\mathrm{T}\right)^{-1} \cdot \text{diag}(\boldsymbol{\mathcal{B}}) \cdot \mathbf{T}, N < L \\ \mathcal{H}(\mathbf{y}) \cdot \left(\frac{\mathbf{I}}{C} + \mathbf{H}^\mathrm{T}\text{diag}(\boldsymbol{\mathcal{B}})\mathbf{H}\right)^{-1} \cdot \mathbf{H}^\mathrm{T} \cdot \text{diag}(\boldsymbol{\mathcal{B}}) \cdot \mathbf{T}, N \geq L \end{cases} \quad (18)$$

Similarly, the kernel version of ECSELM can also be introduced as (6). In the testing process of multi-class classification, one can then declare the predicted label of test instance $\mathbf{y}$ as

$$\hat{c} = \arg\max_{c \in \{1,\cdots,k\}}\{\mathbf{z} \in \mathbb{R}^k | \mathbf{z} = \mathcal{H}(\mathbf{y}) \cdot \boldsymbol{\beta}_{\boldsymbol{\mathcal{B}}}\}_c \quad (19)$$

where $k$ denotes the number of classes.

Notably, it can be figured out from (15) and (18) that the output weight and the final decision have dependency on the cost information vector $\boldsymbol{\mathcal{B}}$ which can be calculated by the weighting matrix $\boldsymbol{\mathcal{W}}$ and the cost matrix $\boldsymbol{\mathcal{M}}$ jointly, and hence, the next step is to solve the cost information vector $\boldsymbol{\mathcal{B}}$ instead of the cost matrix and the weighting matrix.

### B. Evolutionary CSELM

The ECSELM introduces evolutionary search into the framework of CSELM for cost matrix optimization. As we mentioned before, the cost matrix is generally determined in an empirical way which may easily lead to poor generalization performance for cost-sensitive tasks. To address this problem, the cost matrix is also at the first time to be automatically optimized by an evolutionary algorithm (EA). On the basis of the CSELM, the ECSELM is to find the optimal cost matrix $\mathcal{M}$ which makes a better prediction through the output weights $\boldsymbol{\beta}_\mathcal{M}$ with respect to $\mathcal{M}$ such that the total loss between the predicted value and the ground truth reaches the minimum as follows

$$\min_\mathcal{M} \sum_i \mathcal{L}\{\mathbf{t}_i, f_{\text{CSELM}}(\mathbf{x}_i, \boldsymbol{\beta}_\mathcal{M})\} \quad (20)$$
s.t. $l_1 \leq \mathcal{M}_{i,j} \leq l_2$, $\mathcal{M}_{i,i} = 0, i = 1, \cdots, N; j = 1, \cdots, N$

where $l_1$ and $l_2$ are the low and upper bounds, $N$ is the number of training samples, $\mathcal{L}$ is the classification or regression loss function, $\mathbf{t}_i$ is the label vector of sample $\mathbf{x}_i$, and $f_{\text{CSELM}}$ denotes the proposed CSELM decision function.

**Algorithm 1. ECSELM**
**Input:** The training set $\{\mathbf{x}_i\}_{i=1}^N$, the training target matrix $\mathbf{T}$.
**Initialize:** the weighting matrix $\mathcal{W}$ and cost matrix $\mathcal{M}$.
**Procedure:**
1. Randomly select input weights $\mathbf{w}$ and hidden biases $\mathbf{b}$.
2. Compute the cost information vector $\mathcal{B}$ with $\mathcal{B}_i = \sum_j (\mathcal{W} \cdot \mathcal{M})_{ij}$.
3. Compute the hidden layer output matrix $\mathbf{H}$ of training set and the feature mapping $\mathcal{H}(\mathbf{y})$ using (16) and (17).
4. Compute the output weights $\boldsymbol{\beta}_\mathcal{B}$ using (15).
5. Obtain the optimal $\mathcal{B}^*$ by solving the optimization problem (21) using Algorithm 2.
6. Compute the optimal output weights $\boldsymbol{\beta}_{\mathcal{B}^*}$ by substituting $\mathcal{B}^*$ to (15).
**Output:** $\boldsymbol{\beta}_{\mathcal{B}^*}$.

However, it can be found that the output weight matrix $\boldsymbol{\beta}$ as shown in (15) is associated with $\mathcal{B}$, which is indeed calculated by multiplying an unknown/known weighted matrix $\mathcal{W}$ with the unknown cost matrix $\mathcal{M}$. For convenience, the optimization problem (20) seeking for $\mathcal{M}$ can thus be intuitively transformed as the following

$$\mathcal{B}^* = \arg\min_\mathcal{B} \sum_i \mathcal{L}\{\mathbf{t}_i, f_{\text{CSELM}}(\mathbf{x}_i, \boldsymbol{\beta}_\mathcal{B})\} \quad (21)$$
s.t. $l'_1 \leq \mathcal{B}_i \leq l'_2$

where $l'_1$ and $l'_2$ are the new bounds.

By solving (21), i.e. the optimization of the CSELM classifier/predictor model in decision level, the optimal output weight matrix $\boldsymbol{\beta}_{\mathcal{B}^*}$ can be obtained simultaneously with respect to the optimal cost information vector $\mathcal{B}^*$.

Then, the predicted output in decision level of test instance $\mathbf{y}$ can be represented as

$$\hat{c} = \arg\max_{c \in \{1, \cdots, k\}} \{\mathbf{z} \in \mathbb{R}^k | \mathbf{z} = \mathcal{H}(\mathbf{y}) \cdot \boldsymbol{\beta}_{\mathcal{B}^*}\} \quad (22)$$

The proposed ECSELM is summarized in **Algorithm 1**.

*C. Optimization*

To find the optimal $\mathcal{B}$, evolutionary algorithm (EA) is employed intuitively under the only boundary constraint. EA is a population based stochastic search strategy that search for near-optimal solutions. EA tries to evolve an individual into a new individual with better fitness by a trial individual, which can be generated using various genetic operators on the raw individuals such that ongoing new effort is made on EA. In this paper, we leverage a new evolutionary algorithm i.e. backtracking search optimization algorithm (BSA) structured in [40] to learn the cost matrix simultaneously. BSA, as a random search method with three basic genetic operators: selection, mutation and crossover used to generate trial individuals, has simple structure such that it is effective, fast and capable of solving multimodal problems. It can be briefly described as four stages in implementation: *initialization*, *selection-I*, *recombination* and *selection-II*. For details, the basic steps of BSA are formulated as follows.

**Algorithm 2. ECS framework**
**Input:** The population size $N$, problem dimension $D$, lower and upper bounds $l_d$ and $l_u$, the maximal iterations *epoch*;
**Procedure:**
1. *Initialization*:
1.1. Population generation $\mathcal{P}_{i,j} \leftarrow \mathrm{U}(l_d^j, l_u^j)$ using (23);
1.2. Objective function evaluation using (24);
**while** *iteration<epoch* **do**
2. *Selection-I*: update step for historical population.
2.1. Historical population $\mathcal{Q}_{i,j} \leftarrow \mathrm{U}(l_d^j, l_u^j)$ using (25);
2.2. Redefine $\mathcal{Q} \leftarrow \mathcal{P}$ using 'if-then' rule in (26) for memory;
2.3. Permute $\mathcal{Q}' \leftarrow permuting(\mathcal{Q})$ by shuffling (27);
3. *Recombination*: update step for solution population.
3.1. Generate crossover mapping matrix using (28);
3.2. Mutate for new population using (29);
3.3. Boundary control with (30);
3.4. Objective function evaluation with the new population using (31);
4. *Selection-II*: update step for new solution population, global minimum and optimal solution.
4.1. Update population using (32);
4.2. Update the global minimum $\mathcal{F}_{best} := \mathcal{F}_{gmin}$ using (33);
4.3. Update the optimal solution using (34)
**end while**
**Output:** $\mathcal{B}^*$.

1) *Initialization*: generation and evaluation of a population $\mathcal{P}$.
$$\mathcal{P}_{i,j} \sim \mathrm{U}(l_d^j, l_u^j), i = 1, \cdots, N; j = 1, \cdots, D \quad (23)$$
$$\mathcal{F}_i = \mathrm{ObjFun}_i(\mathcal{P}_i), i = 1, \cdots, N \quad (24)$$

where $\mathcal{P}$ is encoded by the solution form of $\mathcal{B}$, $N$ and $D$ denote the population size and problem dimension, $l_d^j$ and $l_u^j$ denote the low and upper bounds with respect to the *j*-th element, U denotes uniform distribution, and $\mathrm{ObjFun}(\cdot)$ denotes the objective function (21).

2) *Selection-I*: update step for historical population $\mathcal{Q}$.
$$\mathcal{Q}_{i,j} \sim \mathrm{U}(l_d^j, l_u^j) \quad (25)$$
$$\text{if } a < b \text{ then } \mathcal{Q} = \mathcal{P}, \forall a, b \sim \mathrm{U}(0,1) \quad (26)$$
$$\mathcal{Q}' = permuting(\mathcal{Q}) \quad (27)$$

where $permuting(\cdot)$ is a random shuffling function, $a$ and $b$ are two random number of uniform distribution. The historical population is for memory characteristics.

3) *Recombination*: update step for solution population $\mathcal{P}'_{new}$.
$$\text{Binary mapping matrix } \mathcal{C}_{N \times D} | 0-1 \quad (28)$$
$$\mathcal{P}_{new} = \mathcal{P} + 3r \cdot \mathcal{C} \odot (\mathcal{Q}' - \mathcal{P}) \quad (29)$$

$$\mathcal{P}'_{new(i,j)} = \begin{cases} l_d^j, & \text{if } rand^1 < rand^2 \text{ and } \mathcal{P}_{new(i,j)} < l_d^j \\ rand \times (l_u^j - l_d^j) + l_d^j, & \text{otherwise} \\ l_u^j, & \text{if } rand^1 < rand^2 \text{ and } \mathcal{P}_{new(i,j)} > l_u^j \\ rand \times (l_u^j - l_d^j) + l_d^j, & \text{otherwise} \end{cases} \quad (30)$$

where $\mathcal{P}'_{new(i,j)}$ represents the *j*-th element of the *i*-th individual, $\odot$ denotes dot product, $r \sim N(0,1)$, $rand^1$ and $rand^2 \sim \mathrm{U}(0,1)$, and $N(0,1)$ denotes standard normal distribution.

Then, evaluate the new population by computing
$$\mathcal{F}'_i = \mathrm{ObjFun}_i(\mathcal{P}'_{new}\{i\}), i = 1, \cdots, N \quad (31)$$
where $\mathcal{P}'_{new}\{i\}$ denotes the *i*-th individual of the population.

4) *Selection-II*: update step for new solution population $\mathcal{P}''_{new}$, global minimum $\mathcal{F}_{gmin}$, and the optimal solution $\mathcal{G}_{opt}$.
$$\mathcal{P}''_{new} = \mathcal{P}'_{new}\{\mathcal{F}'_i < \mathcal{F}_i\} \cup \mathcal{P}\{\mathcal{F}'_i \geq \mathcal{F}_i\}, i = 1, \cdots, N \quad (32)$$
$$\mathcal{F}_{gmin} = \min\{\mathcal{F}\{\mathcal{F}'_i \geq \mathcal{F}_i\} \cup \mathcal{F}'\{\mathcal{F}'_i < \mathcal{F}_i\}\}, i = 1, \cdots, N \quad (33)$$

$\mathcal{G}_{opt} = \mathcal{P}''_{new}\{ind_{opt}|ind_{opt} = \arg_i \min\{\mathcal{F}\{\mathcal{F}'_i \geq \mathcal{F}_i\} \cup \mathcal{F}'\{\mathcal{F}'_i < \mathcal{F}_i\}\}\}$ (34)

where $ind_{opt}$ denotes the index of the optimal individual. Specifically, the proposed ECS framework for problem solution is summarized in **Algorithm 2**.

## IV. HUMAN BEAUTY DATA ANALYSIS

Human beauty analysis is an emerging subject in computer vision and biometric community. Ancient Greek scholars measure the vertical and horizontal distances among eyes, nose, mouth, etc. and propose some general rules such as golden ratio to evaluate the attractiveness of faces. Facial attractiveness assessment using geometric and appearance based features coupled with pattern recognition techniques have been studied separately [41]-[44]. We explore human beauty analysis in this paper because it is recognized as a cost-sensitive learning task [45], and therefore used to evaluate the proposed method.

Recently, a public multi-modality beauty ($M^2B$) database which includes three sub datasets: facial images, dressing images and vocal data, of female persons from eastern and western cultural races have been released online for human beauty study [46], [47]. In this section, we will exploit the proposed ECSELM method on the $M^2B$ database for facial, dressing and vocal attractiveness assessment.

### A. $M^2B$ Database

In $M^2B$ database, the facial, dressing and vocal features were from 620 eastern females (i.e. Chinese, Korean and Japanese) and 620 western females (i.e. Caucasian, consisting of Angles, Celtic, Latin and Germanic). For facial beauty analysis, geometric (denoted as "G") and appearance (denoted as "A") based features were studied separately. The specific details of facial, dressing and vocal feature extraction methods and the attractiveness score acquisition in different modality can be found in [46]. The facial, dressing and vocal features with 300, 300 and 50 dimensions after PCA reduction were used. Some examples of facial images of eastern and western females with landmark points and some examples of dressing images have been shown in Fig.1 and Fig.2, respectively. We observe from Fig.1 that the facial images in $M^2B$ database contain abrupt features such as illumination, poses, occlusions, and expressions. These features also contribute to facial attractiveness while in existing work only fontal faces with restricted setting were used in facial beauty analysis. The attractiveness scores (ground truth) of facial, dressing and vocal features for each person were normalized within [1, 10] from $k$-wise ratings of raters [46].

### B. Parameters Setting

In experiments, two parameters $L$ and $C$ are involved in ECSELM. The number $L$ of hidden neurons is selected from 100 to 500, and the penalty parameter $C$ is selected from $2^0$ to $2^{30}$. The parameter sensitivity of the algorithms are explored in Section VIII by changing the $C$ value and the number $L$ for presenting the best results. In optimization, both the maximum population size and the search epochs are set as 100, the lower and upper boundary is set as -1 and 1, respectively. Notably, the population size and epochs can be accordingly adjusted.

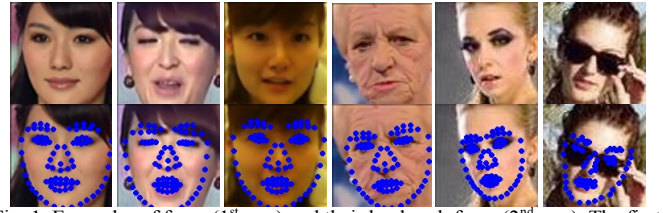

Fig. 1. Examples of faces (1st row) and their landmark faces (2nd row). The first 3 faces in each row denote eastern females, and the last 3 faces are from western

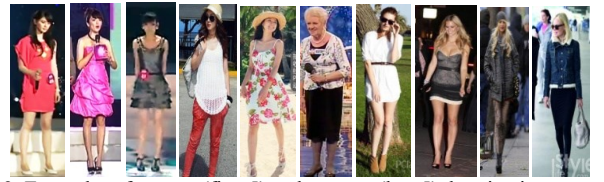

Fig. 2. Examples of eastern (first 5) and western (last 5) dressing images

### C. Attractiveness Assessment: Beauty Recognition

To qualitatively evaluate the beauty, the raw attractiveness scores within [1, 10] for facial, dressing and vocal features have been divided into five levels of 1 (1~2), 2 (2~4), 3 (4~6), 4 (6~8), and 5 (8~10) which correspond to the beauty quality of "poor", "fair", "good", "very good", and "excellent", respectively. In experiment, the attractiveness assessment of eastern (denoted as "E") and western (denoted as "W") females is studied separately. 400 females are randomly selected as training set, and the remaining 220 females are determined as testing set. Then, we run each procedure 10 times under each experiment, and the average rank-1 recognition accuracy (i.e. the ratio between the number of correctly recognized samples and the number of total testing samples) and the standard deviation for each method have been provided. The compared methods are divided into three categories:

- The comparisons with ELM based methods including basic ELM, KELM, $W^1$ELM, $W^2$ELM and E-ELM are explored.
- The comparisons with subspace methods and their cost-sensitive extensions including CSPCA, CSLDA, CSLPP, and CSMFA are presented.
- The comparisons with generic classifiers including $k$-nearest neighbors (kNN), support vector machine (SVM) and least square SVM (LSSVM), are provided. We also compare with cost-interval SVM (CISVM) [38] which was first proposed for addressing the cost-sensitive matrix problem using cost interval. Additionally, we have compared a cost-sensitive ordinal regression (CSOR) [45] that is used for facial beauty.

The rank-1 recognition results of human attractiveness using ELM based methods are presented in Table I, from which we find that the recognition rate obtained by ECSELM for each task is about 10% higher than other ELMs. The appearance based features ('A') outperforms geometric feature ('G') in attractiveness assessment. The reason may be that the faces contain different types of abrupt features such as illumination, poses, color, texture, etc. The results of cost-sensitive subspace methods e.g. CSPCA, CSLDA, CSLPP, and CSMFA, are shown in Table II, from which we can find that ECSELM still outperforms other subspace learning methods.

Table III presents the comparisons with the generic classifiers (e.g. kNN, SVM and LSSVM) and two cost-sensitive methods (e.g. CISVM and CSOR). The number of nearest neighbors is empirically set as 30. We can observe:

TABLE I
RANK-1 RECOGNITION ACCURACY (%) OF FACIAL, DRESSING, AND VOCAL ATTRACTIVENESS USING ELM BASED METHODS

| Feature | Race | ELM | KELM | W$^1$ELM | W$^2$ELM | E-ELM | ECSELM |
|---|---|---|---|---|---|---|---|
| Facial (G) | E | 32.91±5.50 | 26.05±3.23 | 19.68±5.42 | 24.91±8.21 | 35.27±13.1 | **48.51±1.30** |
| | W | 33.50±4.34 | 26.36±5.83 | 19.55±7.39 | 23.64±6.41 | 39.18±3.67 | **49.36±1.61** |
| Facial (A) | E | 33.45±3.89 | 36.36±8.72 | 21.91±9.54 | 24.00±6.98 | 38.91±6.01 | **50.45±0.81** |
| | W | 35.23±10.5 | 34.14±7.70 | 21.32±9.32 | 23.77±2.11 | 40.55±5.56 | **52.27±0.92** |
| Dressing | E | 37.05±5.80 | 39.68±14.5 | 22.91±5.09 | 28.09±8.09 | 43.45±6.30 | **55.45±1.21** |
| | W | 30.09±3.01 | 35.09±3.71 | 17.27±6.90 | 23.55±9.26 | 37.27±4.64 | **47.27±0.98** |
| Vocal | E | 44.05±7.08 | 41.41±2.49 | 27.91±6.41 | 30.91±5.79 | 48.35±5.82 | **56.18±0.93** |
| | W | 37.14±2.94 | 35.73±7.41 | 21.73±9.70 | 25.45±5.49 | 40.82±6.35 | **54.76±1.13** |

TABLE II
RANK-1 RECOGNITION ACCURACY (%) OF FACIAL, DRESSING, AND VOCAL ATTRACTIVENESS USING SUBSPACE BASED METHODS

| Feature | Race | PCA-NN | CSPCA-NN | LDA-NN | CSLDA-NN | LPP-NN | CSLPP-NN | CSMFA-NN | ECSELM |
|---|---|---|---|---|---|---|---|---|---|
| Facial (G) | E | 29.50±1.85 | 30.27±1.96 | 27.45±3.59 | 28.64±2.47 | 28.86±2.28 | 29.90±3.05 | 29.09±3.51 | **48.51±1.30** |
| | W | 29.23±2.64 | 29.23±2.41 | 30.22±2.77 | 30.50±2.51 | 30.77±3.91 | 29.50±2.49 | 29.41±1.91 | **49.36±1.61** |
| Facial (A) | E | 31.41±2.38 | 32.68±2.22 | 28.45±0.84 | 29.41±2.19 | 30.36±2.56 | 30.41±2.75 | 31.82±2.48 | **50.45±0.81** |
| | W | 28.55±2.94 | 28.55±2.26 | 26.09±2.99 | 26.59±2.44 | 28.91±2.34 | 28.00±3.28 | 31.50±2.85 | **52.27±0.92** |
| Dressing | E | 35.45±3.60 | 30.41±2.78 | 32.14±2.17 | 33.14±2.83 | 38.68±2.96 | 39.95±1.47 | 37.86±4.48 | **55.45±1.21** |
| | W | 27.27±3.72 | 24.82±2.64 | 23.55±2.12 | 24.41±2.45 | 23.73±4.18 | 29.09±3.34 | 29.82±2.47 | **47.27±0.98** |
| Vocal | E | 37.77±2.25 | 38.36±3.86 | 39.77±3.46 | 39.05±2.81 | 43.09±3.32 | 41.59±3.12 | 40.63±2.67 | **56.18±0.93** |
| | W | 33.14±3.50 | 33.59±3.03 | 34.41±2.80 | 33.91±4.33 | 34.68±2.14 | 34.50±2.37 | 36.77±3.07 | **54.76±1.13** |

TABLE III
RANK-1 RECOGNITION ACCURACY (%) OF FACIAL, DRESSING, AND VOCAL ATTRACTIVENESS USING BASELINE CLASSIFIERS

| Attribute | Race | KNN | SVM | LSSVM | CISVM | CSOR | ECSELM |
|---|---|---|---|---|---|---|---|
| Facial (A) | E | 36.45±3.03 | 36.59±1.59 | 36.23±2.17 | 34.91±2.79 | 39.61±1.19 | **50.45±0.81** |
| | W | 37.64±3.57 | 38.64±2.37 | 39.59±3.11 | 37.20±3.01 | 42.78±1.64 | **52.27±0.92** |
| Dressing | E | 41.13±3.22 | 39.82±3.12 | 40.68±2.39 | 39.59±5.23 | 42.20±2.03 | **55.45±1.21** |
| | W | 35.68±2.35 | 36.18±2.05 | 33.50±2.75 | 35.45±3.17 | 35.71±2.15 | **47.27±0.98** |
| Vocal | E | 43.30±2.97 | 46.59±1.89 | 47.27±3.08 | 45.50±3.15 | 49.95±2.02 | **56.18±0.93** |
| | W | 38.23±3.90 | 39.05±3.25 | 37.18±3.09 | 37.82±3.70 | 40.67±2.30 | **54.76±1.13** |

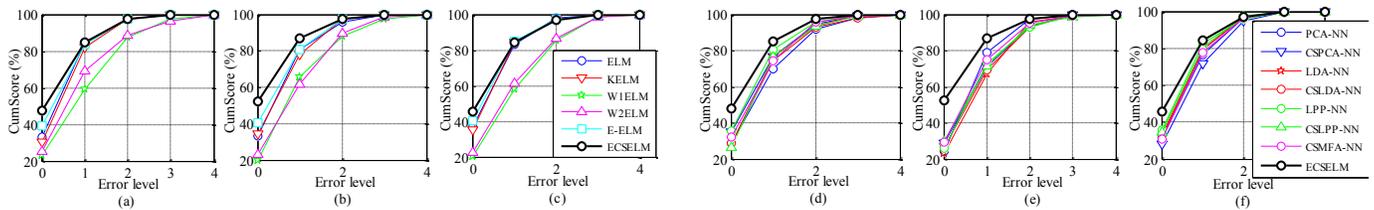

Fig. 3. Cumulative scores of facial attractiveness recognition using ELM based methods (a~c) and subspace based methods (d~f). Eastern: (a) and (d); Western: (b) and (e); Eastern+Western: (c) and (f)

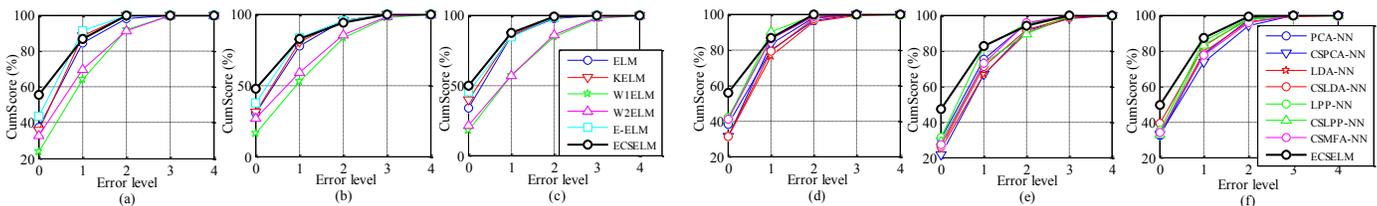

Fig. 4. Cumulative scores of dressing attractiveness recognition using ELM based methods (a~c) and subspace based methods (d~f). Eastern: (a) and (d); Western: (b) and (e); Eastern+Western: (c) and (f)

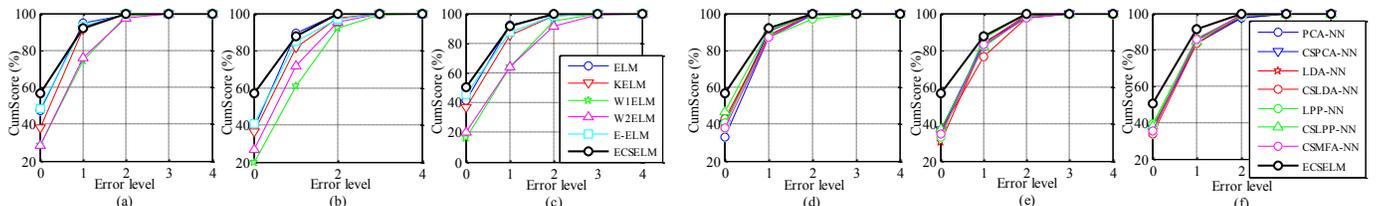

Fig. 5. Cumulative scores of vocal attractiveness recognition using ELM based methods (a~c) and subspace based methods (d~f). Eastern: (a) and (d); Western: (b) and (e); Eastern+Western: (c) and (f)

- For different tasks, CISVM seems to be worse than other methods. The reason may be that CISVM tends to address the problem of cost matrix using cost interval (CI) instead of a precise cost value, but CI is still predefined and task-dependent. In addition, the training complexity of SVM increases, depending on the specific size of the cost interval.
- Though CSOR is improved compared with SVM by introducing cost-sensitive element, the cost matrix construction is prior defined and lack of flexible property for different tasks and new environments.
- ECSELM performs the best recognition accuracy with an approximate 10% improvement.

In attractiveness assessment of 5 levels, the cumulative score, measured in recognition [48]-[52], is also used to evaluate the proposed methods. The cumulative score can be defined as

$$CumScore(\vartheta) = N_{e \leq \vartheta}/N_{test} \times 100\% \quad (35)$$

where $\vartheta$ denotes the tolerated error level, $N_{e \leq \vartheta}$ denotes the number of testing instances whose absolute error $e$ between the predicted label and the true label less than $\vartheta$ ( $\vartheta = 0, 1, 2, \cdots, k-1$ ). $N_{test}$ denotes the number of total testing instances and $k$ is the class number. $CumScore(0)$ denotes the rank-1 recognition. The $CumScore$ curves by using ELM and subspace based methods have been illustrated in Fig.3, Fig.4, and Fig. 5, from which we can see that the proposed ECSELM shows the best performance. Besides, the attractiveness score estimation and further comparisons with the nearest neighbor (NN), ridge regression, neural network, dual-supervised feature-attribute-task (DFAT) [46], and latent DFAT (LDFAT) [47] methods are exploited by strictly following [46] with a standard 2-fold cross validation test in experiments. The cross-validation process is repeated 10 times and the average value is presented to be the final results. In estimation of the attractiveness scores which is scaled within [1, 10], the mean absolute error (*MAE*) defined as $MAE = \sum_{i=1}^{N_{test}} |\hat{y}_i - t_i|/N_{test}$ is used for performance measurement and comparison, where $N_{test}$ is the number of test instances, $\hat{y}_i$ and $t_i$ are the estimated score and the ground truth of instance *i*, respectively.

The results of facial, dressing, and vocal attractiveness score estimation are shown in Table IV. Some results other than ELM methods are simply copied from [46], [47]. The proposed ECSELM shows a competitive performance by comparing with state-of-the-art LDFAT. Comparatively, vocal attractiveness score prediction is better than dressing and facial attractiveness prediction. To study the aesthetic difference between cultures or races, we have conducted the cross culture experiment, that is, we learn a model from the one culture and tests on the other culture, denoted as E→W and W→E, alternatively. The results of between-culture are shown in Table V, from which, we can find that ECSELM shows the lowest MAE for prediction.

## V. FACE DATA ANALYSIS

In this section, we conduct face recognition and face verification experiments using the proposed methods. This section aims at testing the usefulness of the proposed methods, whilst the comparisons with those face recognition methods are not concentrated because this work is not specifically presented for face recognition. We test on two benchmark face datasets: AR face database [50] that contains the faces of 100 persons (50 males and 50 females) and the challenging LFW (labeled faces in the wild) [53] that consists of 13,233 images of 5749 people in unrestricted environments.

### A. Experiment on AR Dataset

We follow the same experimental setting as [52] in which 7

TABLE IV
MAE OF ATTRACTIVENESS SCORES ESTIMATION

| Method | Facial | | Dressing | | Vocal | |
|---|---|---|---|---|---|---|
| | E | W | E | W | E | W |
| NN | 2.10 | 1.91 | 1.50 | 2.02 | 1.39 | 1.78 |
| Ridge Regression | 1.89 | 1.83 | 1.39 | 1.76 | 1.15 | 1.37 |
| Neural Network | 1.82 | 1.75 | 1.37 | 1.62 | 1.12 | 1.38 |
| DFAT | 1.52 | 1.48 | 1.26 | 1.46 | 1.01 | 1.24 |
| LDFAT | 1.46 | 1.46 | **1.14** | 1.37 | 0.96 | 1.14 |
| ELM | 1.55 | 1.53 | 1.29 | 1.56 | 1.04 | 1.27 |
| KELM | 1.72 | 1.52 | 1.32 | 1.51 | 1.33 | 1.77 |
| $W^1$ELM | 1.82 | 1.79 | 1.56 | 1.69 | 1.54 | 1.61 |
| $W^2$ELM | 1.71 | 1.76 | 1.50 | 1.69 | 1.45 | 1.58 |
| E-ELM | 1.46 | 1.45 | 1.21 | 1.48 | **0.95** | 1.21 |
| ECSELM | **1.40** | **1.43** | **1.14** | **1.36** | 0.97 | **1.13** |

TABLE V
MAE OF CROSS-CULTURE ATTRACTIVENESS ESTIMATION

| Method | Facial | | Dressing | | Vocal | |
|---|---|---|---|---|---|---|
| | E→W | W→E | E→W | W→E | E→W | W→E |
| DFAT | 1.91 | 2.22 | 2.55 | 2.71 | 1.55 | 1.62 |
| LDFAT | 1.57 | **1.43** | 1.61 | 1.40 | 1.44 | 1.32 |
| ELM | 1.52 | 1.56 | 1.50 | 1.29 | 1.27 | 1.03 |
| KELM | 1.57 | 1.53 | 1.62 | 1.52 | 1.76 | 1.78 |
| $W^1$ELM | 1.74 | 1.80 | 1.73 | 1.72 | 1.75 | 1.60 |
| $W^2$ELM | 1.66 | 1.71 | 1.70 | 1.66 | 1.72 | 1.56 |
| E-ELM | 1.53 | 1.48 | 1.51 | 1.27 | **1.22** | 1.07 |
| ECSELM | **1.47** | 1.51 | **1.46** | **1.25** | 1.23 | **0.99** |

facial images per person from session 1 with illumination and expression changes were used for training and the other 7 images per person with the same condition from session 2 were used for testing. Eigenface [54] with 300 dimensions after PCA is used as feature in experiment. For fair comparisons, we follow the same train/test split for all methods.

We have compared the proposed evolutionary cost-sensitive methods with generic classifiers such as nearest neighbor (NN), nearest subspace (NS) and linear SVM, cost-sensitive subspace analysis based methods (e.g. CSPCA, CSLPP, CSMFA, and CSLDA), and ELM based methods (e.g. ELM, KELM, WELM and E-ELM). In addition, three specialized cost sensitive face recognition methods including multiclass cost-sensitive kNN (mckNN) [29], multiclass cost-sensitive SVM (mcSVM) [55], and multiclass cost-sensitive kernel logistic regression (mcKLR) [29] are also compared in this paper. The kernel case of ECSELM is considered in face recognition (FR) application. Some baseline results are from literature [52].

The results of the ELM with penalty coefficient $C=2^5$ and subspace based methods are presented in Table VI, from which we have the following observations:
- KELM shows an obvious superiority (87.1%) in recognition compared with conventional ELM (81.9%) and WELM (82.7%). More significantly, the proposed CSELM performs a recognition rate of 89.4% with 2.3% improvement compared with KELM, which clearly demonstrates the effect of cost-sensitive learning in ELM.
- In the subspace based methods, CSLPP performs the worst. The possible reason is that the characteristic of low dimensional embedding in manifold with LPP is not dominant in AR database, and make the learned projection fail. Compared with CSPCA and CSMFA, CSLDA shows much better performance due to its discriminative ability.

TABLE VI
RECOGNITION RATE (%) OF ELM AND COST-SENSITIVE SUBSPACE BASED METHODS

| Methods | CSPCA | CSLPP | CSMFA | CSLDA | ELM | KELM | WELM | E-ELM | CSELM |
|---|---|---|---|---|---|---|---|---|---|
| Recognition rate (%) | 68.8 | 45.5 | 69.5 | 86.4 | 81.9 | 87.1 | 82.7 | 86.6 | **89.4** |

TABLE VII
COMPARISONS WITH BASELINES AND STATE-OF-THE-ART COST-SENSITIVE FACE RECOGNITIONS

| Methods | NN | NS | SVM | CISVM | mckNN | mcSVM | mcKLR | ECSELM |
|---|---|---|---|---|---|---|---|---|
| Recognition rate (%) | 71.4 | 76.0 | 87.1 | - | 83.2 | 86.6 | 92.2 | **92.7** |

- ELMs show better flexibility and competitiveness in recognition than subspace methods.

To evaluate our ECSELM methods, we present the results of several popular classifiers and three cost-sensitive face recognition methods in Table VII, from which we have following observations:

- The cost-sensitive face recognition methods (e.g. mckNN, mcSVM and mcKLR) outperform the conventional classifiers, with 92.2% recognition rate obtained by mcKLR as a kernel logistic regression. Comparatively, mcSVM obtains an inferior performance (86.6%).
- The proposed ECSELM method shows the best recognition performance (92.7%) among all the existing methods presented in this paper. Compared with CSELM in Table VI, a further improvement of 3.3% recognition accuracy is obtained. The superior performance demonstrates that the proposed evolutionary cost-sensitive learning in this paper can effectively improve face recognition. Another merit of ECSELM is that, it can predict the label of a given instance intuitively without using multi-class voting mechanism addressed in SVM.

Note that the result of CISVM is not given because there is no report for its use in face recognition. With rigorous consideration, we have downloaded their released codes of CISVM and run the codes on AR data. The obtained recognition accuracy is approximately 28%. Furthermore, the *CumScore* curves with error level $\vartheta$ changes from 0 to 99 (100 classes in AR) are described in Fig.6, which clearly demonstrates that the proposed CSELM and ECSELM outperform other methods for face recognition.

### B. Experiment on LFW Dataset

In this section, we evaluate our methods on the LFW dataset which is commonly regarded as a challenging dataset for unrestricted face verification and matching in the wild, since the faces taken from *Yahoo! News* show large variations in pose, illumination, expression, age, etc. Two pairs of faces are shown in Fig.7. The dataset is organized into two views:

- In *view* 1, a set consisting of 2200 pairs for training and 1000 pairs for testing is developed for model selection.
- In *view* 2, 6000 pairs for 10-fold cross-validation are developed. In each fold, 600 pairs with 300 similar pairs and 300 dissimilar pairs are contained.

Note that the experimental setup for face verification is different from the standard face recognition that fair pairs are given and the decision on each pair is generally made as "same" (positive pair) or "not same" (negative pair) without knowing the identity information of each person.

For this dataset, state-of-the-art metric learning methods [56],

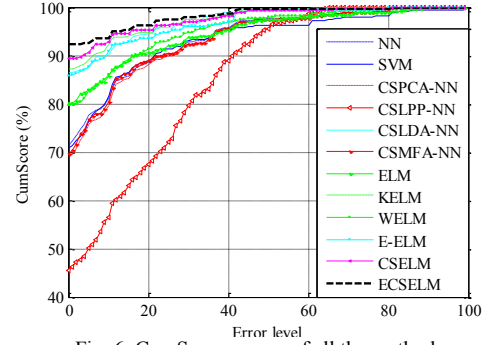
Fig. 6. CumScore curves of all the methods

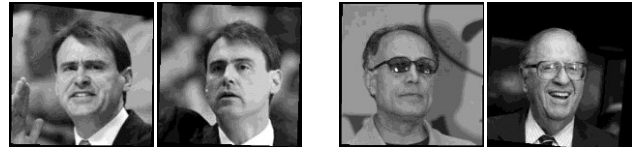
Fig. 7. Sample images of one "same" pair and one "not same" pair from LFW.

[57] are generally explored over intra-personal subspace instead of the generic classifiers (e.g. SVM). To make the proposed methods applicable in LFW, the feature vector that can reflect the similarity information is set for each pair. We do the experiments by following the standard protocol of LFW, and the experimental setup is presented as follows.

For face feature extraction, two kinds of feature descriptors, i.e. local binary patterns (LBP) and scale invariant feature transformation (SIFT) are used, respectively. Each face is then represented as a 300-dimensional vector after PCA [56]. Due to the lack of full class label information, for evaluating the proposed methods in this scenario, we represent a face pair using five similarity metrics: correlation coefficient, Euclidean distance, cosine distance, Mahalanobis distance and bilinear similarity function with positive semi-definite (p.s.d.) matrix learned in [56]. Hence, a 5-dimensional vector is formulated to represent each similar/dissimilar pair, and a binary classifier is trained using our proposed methods. Following the 10-fold cross-validation protocol for performance evaluation on *view* 2, the mean verification accuracies of 10 folds are reported.

The results of ELMs and cost-sensitive subspace methods are reported in Table VIII, from which we can observe that

- ELM based methods outperform the subspace methods regularly with similar effect in AR experiments. Nevertheless, the standard deviations of ELMs are higher than others. The possible reason is that the hidden layer output matrix of ELM is activated with randomly generated weights and bias.
- CSMFA shows the worst face verification performance among all the methods. The possible reason is that the constructed locality graph using $k$ nearest neighbors of each input sample fails on the LFW database consisting of many *face pairs*, such that the intra-sample information is lost.
- The proposed ECSELM outperforms other methods by comparing with cost-sensitive subspace methods and conventional ELM methods.

TABLE VIII
RECOGNITION RATE (%) OF ELM AND COST-SENSITIVE SUBSPACE BASED METHODS

| Method | CSPCA | CSLDA | CSLPP | CSMFA | ELM | KELM | WELM | E-ELM | ECSELM |
|---|---|---|---|---|---|---|---|---|---|
| LBP descriptor | 82.87±1.18 | 82.45±1.69 | 84.30±1.45 | 53.18±1.70 | 85.40±2.81 | 85.72±3.07 | 85.93±2.24 | 86.97±3.10 | **87.97±1.37** |
| SIFT descriptor | 78.65±1.14 | 79.27±1.23 | 81.65±1.74 | 52.76±1.35 | 83.40±1.43 | 84.37±2.55 | 84.85±1.35 | 83.77±2.51 | **86.60±1.25** |

TABLE IX
RECOGNITION RATE (%) COMPARISONS WITH STATE-OF-THE-ART METRIC LEARNING METHODS ON LFW

| Method | SILD | ITML | LDML | CSML | KISSME | DML-eig | LMLML | SubSML | ECSELM |
|---|---|---|---|---|---|---|---|---|---|
| LBP descriptor | 80.07±4.27 | 83.98±1.52 | 82.27±1.83 | 85.57±1.64 | 83.37±1.71 | 82.28±1.30 | 86.13±1.68 | 86.73±1.68 | **87.97±1.37** |
| SIFT descriptor | 80.85±1.93 | 81.45±1.45 | 81.05±1.52 | - | 83.08±1.77 | 81.27±7.27 | - | 85.55±1.93 | **86.60±1.25** |

TABLE X
RANK-1 RECOGNITION OF GASES USING ELM BASED METHODS AND SUBSPACE ANALYSIS BASED NN CLASSIFIERS

| Method | PCA | CS-PCA | LDA | CS-LDA | LPP | CS-LPP | CS-MFA | ELM | KELM | $W^1$ELM | $W^2$ELM | E-ELM | ECS-ELM |
|---|---|---|---|---|---|---|---|---|---|---|---|---|---|
| HCHO | 95.24 | 95.24 | 92.06 | 92.06 | 92.06 | 93.65 | 95.24 | 96.83 | 88.89 | 80.95 | 90.47 | 96.83 | 98.41 |
| $C_6H_6$ | 87.50 | 87.50 | 83.33 | 79.17 | 91.67 | 91.67 | 87.50 | 83.22 | 91.67 | 95.83 | 95.83 | 91.67 | 100.0 |
| $C_7H_8$ | 100.0 | 100.0 | 100.0 | 100.0 | 100.0 | 100.0 | 100.0 | 95.45 | 95.45 | 100.0 | 100.0 | 95.45 | 100.0 |
| CO | 95.00 | 95.00 | 70.00 | 85.00 | 85.00 | 85.00 | 95.00 | 90.00 | 100.0 | 100.0 | 95.00 | 90.00 | 100.0 |
| $NH_3$ | 100.0 | 100.0 | 100.0 | 100.0 | 95.00 | 95.00 | 100.0 | 95.00 | 95.00 | 100.0 | 95.00 | 95.00 | 100.0 |
| $NO_2$ | 84.62 | 84.62 | 69.23 | 69.23 | 76.92 | 76.92 | 84.62 | 61.53 | 76.92 | 84.62 | 84.62 | 76.92 | 84.62 |
| *ARR* | 93.73 | 93.73 | 85.77 | 87.58 | 90.11 | 90.37 | 93.73 | 87.01 | 91.32 | 93.57 | 93.48 | 90.98 | **97.17** |
| *TRR* | 94.44 | 94.44 | 88.27 | 89.51 | 91.35 | 91.98 | 94.44 | 90.74 | 91.36 | 90.74 | 93.21 | 93.21 | **98.15** |

TABLE XI
RANK-1 RECOGNITION RATE (%) OF GASES USING BASELINES AND GENERAL CLASSIFIERS FOR E-NOSE

| Method | SVM | PCA-SVM | KSVM | LDA | PCA-LDA | PLS-DA | KLDA | KPLS-DA | CISVM | CSELM | ECSELM |
|---|---|---|---|---|---|---|---|---|---|---|---|
| HCHO | 98.41 | 98.41 | 98.41 | 88.89 | 82.54 | 93.65 | 95.24 | 98.41 | 93.65 | 92.06 | 98.41 |
| $C_6H_6$ | 79.17 | 91.67 | 87.50 | 66.67 | 58.33 | 45.83 | 100.0 | 91.67 | 83.33 | 95.83 | 100.0 |
| $C_7H_8$ | 100.0 | 100.0 | 100.0 | 90.91 | 86.36 | 68.18 | 95.45 | 95.45 | 72.73 | 100.0 | 100.0 |
| CO | 100.0 | 65.00 | 100.0 | 100.0 | 90.00 | 75.00 | 95.00 | 95.00 | 80.00 | 100.0 | 100.0 |
| $NH_3$ | 90.00 | 100.0 | 95.00 | 90.00 | 90.00 | 70.00 | 95.00 | 90.00 | 95.00 | 100.0 | 100.0 |
| $NO_2$ | 69.23 | 30.77 | 76.92 | 30.77 | 30.77 | 23.08 | 76.92 | 69.23 | 30.77 | 84.62 | 84.62 |
| *ARR* | 89.47 | 80.97 | 92.97 | 77.87 | 73.00 | 62.62 | 92.94 | 89.96 | 75.91 | 95.41 | **97.17** |
| *TRR* | 92.59 | 88.27 | 95.06 | 82.72 | 77.16 | 72.22 | 94.44 | 93.21 | 82.72 | 95.06 | **98.15** |

Further, we compare our ECSELM with several state-of-the-art metric learning methods such as side information based linear discriminant analysis (SILD) [58], keep it simple and straightforward metric learning (KISSME) [59], cosine similarity metric learning (CSML) [60], information theoretic metric learning (ITML) [61], logistic discriminant metric learning (LDML) [62], distance metric learning with eigenvalue (DML-eig) [63], large margin local metric learning (LMLML) [57] and similarity metric learning over subspace (SubSML) [56], which have been well tested on LFW. The comparison results are shown in Table IX, from which we have following observations:
- Among the metric learning methods, SubSML shows the best performance on both feature descriptors, which reflects the effect of Mahalanobis distance metric and bilinear function in SubSML. Notably the results of CSML and LMLML on SIFT are not given, because they were not reported in [57], [60].
- Our proposed ECSELM performs significantly the best recognition among the state-of-the-art metric learning methods for both descriptors. Besides, a new prospective that group metrics can be integrated as input features for face verification by learning a binary classifier.

## VI. E-NOSE DATA ANALYSIS

E-NOSE is a multi-sensor system comprised of a sensor array with partial specificity coupled with pattern recognition algorithm [64], which can also be recognized as cost-sensitive problem. In this section, we will explore the proposed methods on E-NOSE database for new application of gases recognition (GR), and validate the generality of the proposed methods in cost-sensitive recognition task. The E-NOSE database is prepared based on six kinds of gases (i.e. 6 classes problem), such as formaldehyde (HCHO), benzene ($C_6H_6$), toluene ($C_7H_8$), carbon monoxide (CO), ammonia ($NH_3$) and nitrogen dioxide ($NO_2$) in [65]-[67]. The number of samples for each gas is 188, 72, 66, 58, 60 and 38, respectively. The steady state response of each sensor is extracted as feature, and a 6-dimensional feature vector is formulated as one sample. Two thirds of samples per class are randomly selected as training set.

The rank-1 recognition of each class, average recognition rate (*ARR*) and the total recognition rate (*TRR*) are computed. Notably, *ARR* is the ratio of the summation of all recognition rates and class number, whilst *TRR* is the ratio between the number of correctly classified samples for all classes and the total number of samples. The comparisons with ELM methods, subspace methods, and existing methods are conducted. The rank-1 recognition results of ELM based methods and subspace based learning methods coupled with the nearest neighbor (NN) classifier are presented in Table X, from which we observe that ECSELM performs the best recognition performance with 97.17% of *ARR* and 98.15% of *TRR*.

TABLE XII
SUMMARIZED RECOGNITION ACCURACY (%) ON MULTIPLE TEST DATA FOR STATISTICAL SIGNIFICANCE TEST

| Test data | CSPCA | CSLDA | CSLPP | CSMFA | ELM | KELM | WELM | E-ELM | ECSELM |
|---|---|---|---|---|---|---|---|---|---|
| M$^2$B data (facial) | 32.68 | 29.41 | 30.41 | 31.82 | 33.45 | 36.36 | 21.91 | 38.91 | **50.45** |
| M$^2$B data (dress) | 30.41 | 33.14 | 39.95 | 37.86 | 37.05 | 39.68 | 22.91 | 43.45 | **55.45** |
| M$^2$B data (vocal) | 38.36 | 39.05 | 41.59 | 40.63 | 44.05 | 41.41 | 27.91 | 48.35 | **56.18** |
| AR data | 68.80 | 45.50 | 69.50 | 86.40 | 81.90 | 87.10 | 82.70 | 86.60 | **92.70** |
| LFW data | 78.65 | 79.27 | 81.65 | 52.76 | 83.40 | 84.37 | 84.85 | 83.77 | **86.60** |
| E-NOSE data | 93.73 | 87.58 | 90.37 | 93.73 | 87.01 | 91.32 | 93.57 | 90.98 | **97.17** |

TABLE XIII
STATISTICAL HYPOTHESIS TEST BY USING T-TEST METHOD OF EIGHT PAIRS OF CLASSIFIERS ON MULTIPLE TESTING DATASETS

| Pairs | <CSPCA,ours> | <CSLDA,ours> | <CSLPP,ours> | <CSMFA,ours> | <ELM,ours> | <KELM,ours> | <WELM,ours> | <E-ELM,ours> |
|---|---|---|---|---|---|---|---|---|
| $p$ | 0.0062 | 0.0162 | 0.0047 | 0.0153 | 0.0030 | 0.0094 | 0.0274 | 0.0029 |
| $H (\alpha=0.01)$ | 1 | 0 | 1 | 0 | 1 | 1 | 0 | 1 |
| $H (\alpha=0.05)$ | 1 | 1 | 1 | 1 | 1 | 1 | 1 | 1 |

For comparison with existing methods in E-NOSE classification, we have conducted the experiments using several popular methods such as SVM, LDA, partial least square-discriminant analysis (PLS-DA), and their kernel extensions, e.g. kernel SVM (KSVM), kernel LDA (KLDA), kernel PLS-DA (KPLS-DA) in Table XI, which also clearly demonstrates that the proposed CSELM and ECSELM methods show the best performance. Additionally, LDA and SVM methods after PCA preprocessing (i.e. PCA-LDA and PCA-SVM) are also compared. Note that the one-against-one (OAO) scheme is used in SVM and LDA based methods.

From a variety of applications the generality of the proposed methods is effectively revealed in preliminary, though more tests in large-scale databases can be done to make an effort on the potential of the proposed methods. From the perspective of algorithm, the complexity, computational cost and the convergence of the proposed approach are optimistic. ELM is popular due to their fast computation and good effectiveness. ECSELM is proposed under an evolutionary cost-sensitive learning framework. Evolutionary algorithms are widely used to solve different types of optimization problems for their rapid search in the whole solution space with heuristic and bio-inspired update strategies [68], [69], but EAs do not guarantee finding the global optimum solution for a problem. However, EA has global exploration in the entire search space and local exploitation abilities to find the best solution near a new solution it has discovered [70], [71]. In this paper, the instinct optimization involves three bio-inspired genetic operators, i.e. mutation, crossover and selection. The optimal or near-optimal solutions of the proposed methods can be obtained with finite iterations and a low computational cost.

## VII. PERFORMANCE EVALUATION OF CLASSIFIERS

### A. ROC, AUC and Confusion Matrix Analysis

The performance of different methods has also been analyzed by using ROC curve, AUC and Confusion matrix on three datasets, such as LFW face data, M$^2$B data and E-NOSE

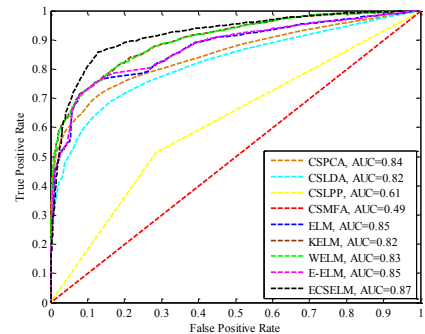
Fig. 8. ROC and AUC analysis on LFW data

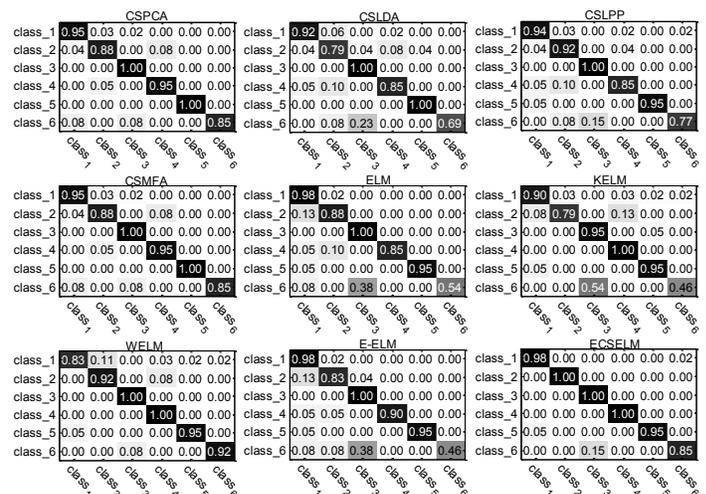
Fig. 9. Confusion matrix analysis based on E-NOSE data

data. LFW data is recognized to be a binary classification task, therefore, ROC and AUC is presented in Fig.8, from which we can observe that the proposed ECSELM method outperforms other methods.

The E-NOSE and M$^2$B data are used as multi-classification tasks, such that the confusion matrix is used for validating the cost-sensitive classification performance, which is shown in Fig. 9 and 10, respectively. The confusion matrix can better show the effectiveness of the proposed method.

### B. Statistical Significance

In this paper, we apply the popular *t-test* and non-parametric *Kruskal-Wallis* method for statistical significance test of 9 different methods on multiple test datasets [72] in a pair-wise manner. The summarized recognition results are shown in Table XII. Two variables $H$ and $p$ are computed using *t-test* on the results from each pair of classifiers, where $p$ denotes the probability of observing the given results, $H=1$ denotes that the null hypothesis is rejected and $H=0$ denotes that the null hypothesis cannot be rejected. The test results are shown in Table XIII, from which we can clearly observe that the proposed ECSELM method statistically outperforms other methods at the significant level $\alpha=5\%$. The Kruskal-Wallis test can also demonstrate the statistical significance of our method.

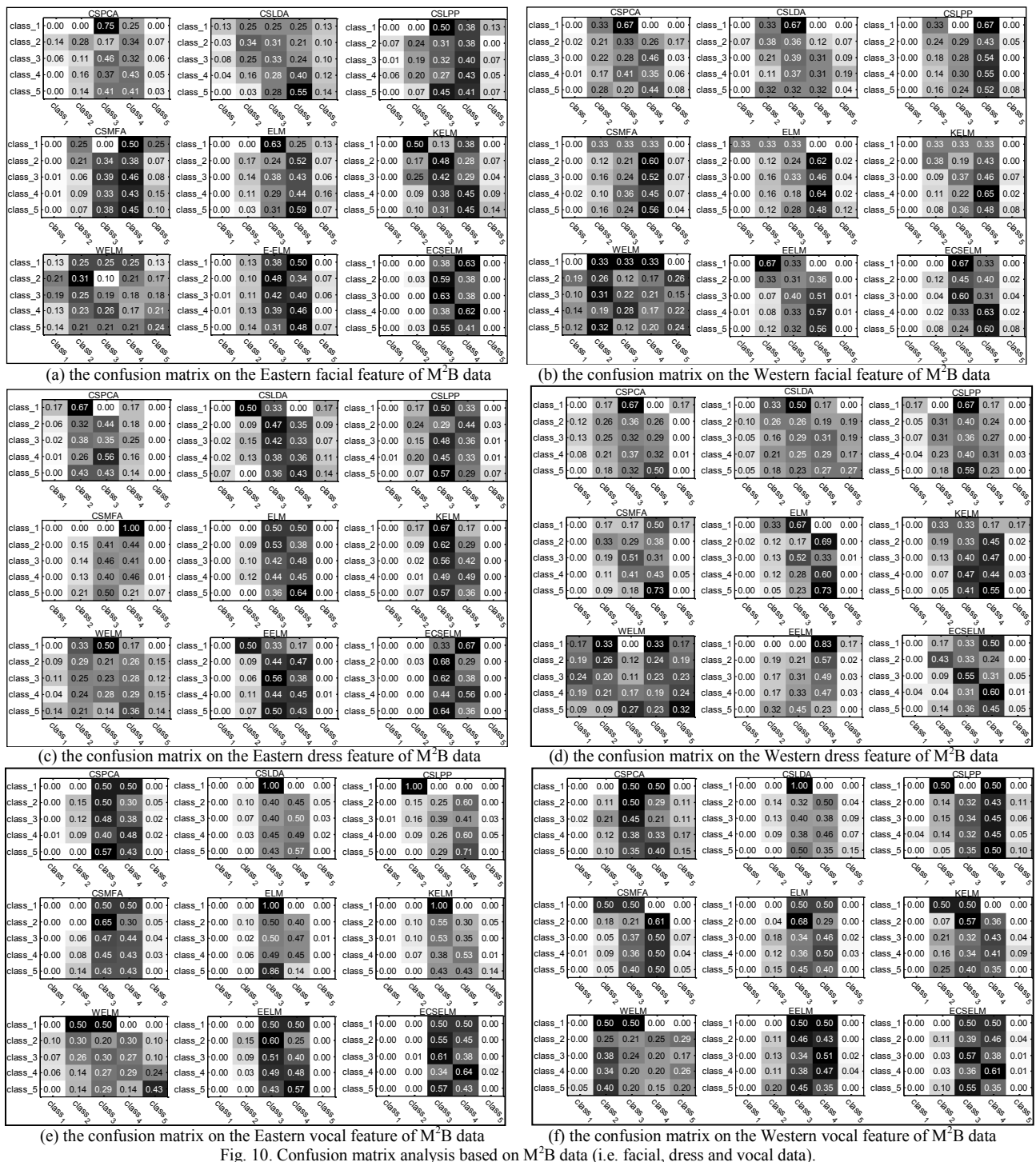

Fig. 10. Confusion matrix analysis based on M²B data (i.e. facial, dress and vocal data).

## VIII. PARAMETER SENSITIVITY ANALYSIS

In the proposed model, there is only one model parameter, i.e. the trade-off coefficient $C$. For different datasets, the parameter variation may show different performance. So, we use different $C$ values from the set $\{2^0, 2^5, 2^{10}, 2^{20}, 2^{30}\}$. Fig.11 shows the performance variations with different penalty coefficient $C$ on FR, LFW and E-NOSE data, from which we see that our method and standard ELM are not sensitive to the trade-off parameter variation, and better performance for AR, LFW, and E-NOSE can be obtained when $C$ is set as $2^5$, $2^{10}$, and $2^{20}$, respectively. Notably, WELM is denoted by $W^2$ELM.

TABLE XIV
TOTAL TRAINING AND TESTING TIME ON LFW DATASET OF ONE FOLD

| Method | CSPCA | CSLDA | CSLPP | CSMFA | ELM | KELM | WELM | E-ELM | ECSELM |
|---|---|---|---|---|---|---|---|---|---|
| Time (s) | 412.08 | 26.39 | 331.77 | 6731.9 | 2.61 | 58.48 | 4.47 | 38.60 | 237.81 |

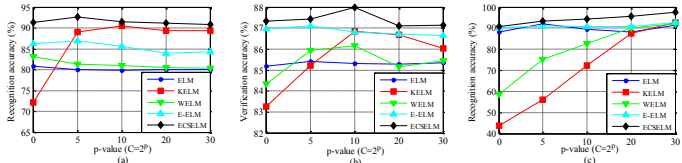

Fig. 11. Performance variation *w.r.t.* the parameter $C=2^p$ in ELM based methods: (a) AR with $L=300$; (b) LFW with $L=100$; (c) E-NOSE with $L=200$.

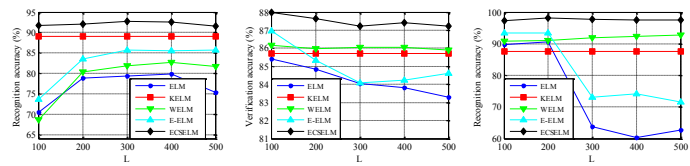

Fig. 12. Performance variation *w.r.t.* the number of hidden neurons $L$ in ELMs: (a) AR with $C=2^5$; (b) LFW with $C=2^{10}$; (c) E-NOSE with $C=2^{20}$.

Additionally, we also studied the performance variation with different number of hidden neurons i.e. $L$. By fixing the best $C$ for each data, we select $L$ from the set {100,200,300,400,500}, and run the ELMs in FR, LFW and E-NOSE data. The results are shown in Fig.12, from which we observe that the there is no large performance variation of the proposed method *w.r.t.* $L$, while the performance of ELM drop dramatically for LFW and E-NOSE data analysis when $L$ is larger than 200. So, the best $L$ for AR, LFW and E-NOSE can be set as 300, 100, and 200, respectively. The hidden layer output matrix of KELM calculated by training samples through a kernel mapping is not associated with $L$, so the recognition rate of KELM is unchanged. We see that the proposed method is more robust to the variation of model parameter and hidden neurons. Note that E-ELM introduces the differential evolutionary method for optimizing the random weights and bias.

## IX. COMPUTATIONAL COMPLEXITY AND TIME ANALYSIS

The proposed algorithms are computationally efficient. For ECSELM, the main steps in Algorithm 1 involve computing the matrix inverse $\left(\frac{I}{c}+diag(\mathcal{B})\mathbf{HH^T}\right)^{-1}$ or $\left(\frac{I}{c}+\mathbf{H^T}diag(\mathcal{B})\mathbf{H}\right)^{-1}$, and the search of the optimal cost information vector $\mathcal{B}$ in Algorithm 2. The hidden layer output matrix $\mathbf{H}$ can be pre-computed. The complexity of matrix multiplication for two matrices of size $m\times n$ and $n\times p$ can be $O(mnp)$. The complexity of Algorithm 2 depends on the population size $N$, problem dimensions $D$ (i.e. the length of vector $\mathcal{B}$), and the number *epochs* of iterations, i.e. $O(N\cdot epochs)$. In the proposed ECSELM, the above matrix computing is included in the loop, i.e. $O(N\cdot epochs\cdot m\cdot n\cdot p)$.

With a naïve Matlab implementation, the algorithms are run on a 2.5GHz Windows machine with 4GB RAM. The computational time based on LFW dataset is presented in Table XIV, from which we observe that:

- KELM and E-ELM needs more computations than ELM and WELM. This is caused by computing the output weights on a higher dimensional kernel matrix and evolutionary search.
- CSPCA and CSLPP cost too much time comparably. For the former, the time is spent on the covariance matrix computation with a large training set. For the latter, a nearest neighbor graph constructed on the training set costs most time. Comparatively, ELMs have much higher computational efficiency than subspace methods.
- The CSMFA cost the most time (6731.9s) among all the methods. The reason is that the time is mostly spent on the computation of the locality graph where $k$ nearest neighbors should be searched for each input vector.
- By inheriting the very low computational complexity of conventional ELM, the proposed ECSELM is faster than cost sensitive subspace methods except the CSLDA.

## X. CONCLUSION AND FUTURE WORK

We have proposed in this paper an evolutionary cost-sensitive extreme learning machine (ECSELM) to address the robustness of ELM in cost-sensitive learning tasks, where different misclassification loss is fully studied. Specifically, the proposed evolutionary cost-sensitive framework is explored for guiding the users to freely and automatically determine the cost matrix that are task specific. To the best of our knowledge, it's the first work to provide a new evolutionary cost-sensitive perspective for ELM. Also, there is no specific approach solving the cost matrix that is commonly defined manually in different scenarios. Extensive experiments have been employed on a variety of application scenarios such as human beauty, face recognition, face verification and E-NOSE. Experimental results and comparisons with several popular methods demonstrate the extremely prominent efficacy and competitive potentials of the proposed approaches for different tasks.

In the future work, it is also challenging to make more insight of extreme learning machines for exploring its deep learning capability, and bring some new perspectives. Additionally, how to improve the evolutionary algorithm by appropriate population generation as indicated in [73] is also motivated. Furthermore, ensemble ELMs may be a good direction, for example, as indicated in recent work [74], a twin ELM framework by integrating two different asymmetric ELMs that are learned with least square and maximum likelihood algorithms respectively, was proposed. More interestingly, as shown in the latest work [75], an idea that the input weights of ELM may not need to be generated randomly was proposed, and proved that they can be replaced with low-discrepancy sequences (LDSs). Therefore, these interesting directions of ELM research can be further explored in the near future.

## ACKNOWLEDGMENT

The authors would like to thank the Associate Editor and anonymous reviewers for their valuable comments, which greatly improved the quality of this paper. This work was supported by National Natural Science Foundation of China (61401048) and Research Fund for Central Universities.

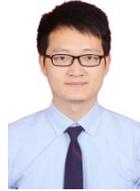

**Lei Zhang** received his Ph.D degree in Circuits and Systems from the College of Communication Engineering, Chongqing University, Chongqing, China, in 2013. He was selected as a Hong Kong Scholar in China in 2013, and worked as a Post-Doctoral Fellow with The Hong Kong Polytechnic University, Hong Kong, from 2013 to 2015. He is currently a Professor/Distinguished Research Fellow with Chongqing University. He has authored more than 40 scientific papers in top journals, including the IEEE TRANSACTIONS ON IMAGE PROCESSING, the IEEE TRANSACTIONS ON NEURAL NETWORKS AND LEARNING SYSTEMS, the IEEE TRANSACTIONS ON MULTIMEDIA, the IEEE TRANSACTIONS ON SYSTEMS, MAN, AND CYBERNETICS: SYSTEMS, the IEEE TRANSACTIONS ON INSTRUMENTATION AND MEASUREMENT, the IEEE SENSORS JOURNAL, INFORMATION FUSION, SENSORS & ACTUATORS B, and ANALYTICA CHIMICA ACTA. His current research interests include machine learning, pattern recognition, computer vision and intelligent systems. Dr. Zhang was a recipient of Outstanding Doctoral Dissertation Award of Chongqing, China, in 2015, Hong Kong Scholar Award in 2014, Academy Award for Youth Innovation of Chongqing University in 2013 and the New Academic Researcher Award for Doctoral Candidates from the Ministry of Education, China, in 2012.

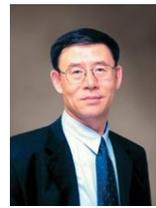

**David Zhang (F'09)** graduated in Computer Science from Peking University in 1974. He received his MSc in 1982 and his PhD in 1985 in Computer Science from the Harbin Institute of Technology (HIT), respectively. From 1986 to 1988 he was a Postdoctoral Fellow at Tsinghua University and then an Associate Professor at the Academia Sinica, Beijing. In 1994 he received his second PhD in Electrical and Computer Engineering from the University of Waterloo, Ontario, Canada. He is a Chair Professor since 2005 at the Hong Kong Polytechnic University where he is the Founding Director of the Biometrics Research Centre (UGC/CRC) supported by the Hong Kong SAR Government in 1998. He also serves as Visiting Chair Professor in Tsinghua University, and Adjunct Professor in Peking University, Shanghai Jiao Tong University, HIT, and the University of Waterloo. He is the Founder and Editor-in-Chief, International Journal of Image and Graphics (IJIG); Book Editor, Springer International Series on Biometrics (KISB); Organizer, the International Conference on Biometrics Authentication (ICBA); Associate Editor of more than ten international journals including IEEE TRANSACTIONS and so on; and the author of more than **10** books, over **300** international journal papers and **30** patents from USA/Japan/HK/China. Professor Zhang is a Croucher Senior Research Fellow, Distinguished Speaker of the IEEE Computer Society, and a Fellow of both IEEE and IAPR.